\documentclass[]{spie}  

\usepackage{amsmath,amsfonts,amssymb}
\usepackage{graphicx}
\usepackage[colorlinks=true, allcolors=blue]{hyperref}
\usepackage{gensymb}
\usepackage [english]{babel}
\usepackage [autostyle, english = american]{csquotes}
\MakeOuterQuote{"}

\title{Limitation of Acyclic Oriented Graphs Matching as Cell Tracking Accuracy Measure when Evaluating Mitosis}

\author[a]{Ye Chen}
\author[a,b]{Yuankai Huo}

\affil[a]{Department of Electrical Engineering \& Computer Science, Vanderbilt University, Nashville, TN, USA 37235}
\affil[b]{Data Science Institute, Vanderbilt University, Nashville, TN, USA 37235}

\authorinfo{Send correspondence to\\Yuankai Huo: E-mail: yuankai.huo@vanderbilt.edu}

\begin{document}
\maketitle
\begin{abstract}
Multi-object tracking (MOT) in computer vision and cell tracking in biomedical image analysis are two similar research fields, whose common aim is to achieve instance level object detection/segmentation and associate such objects across different video frames. However, one major difference between these two tasks is that cell tracking also aim to detect mitosis (cell division), which is typically not considered in MOT tasks. Therefore, the acyclic oriented graphs matching (AOGM) has been used as de facto standard evaluation metrics for cell tracking, rather than directly using the evaluation metrics in computer vision, such as multiple object tracking accuracy (MOTA), ID Switches (IDS), ID F1 Score (IDF1) etc. However, based on our experiments, we realized that AOGM did not always function as expected for mitosis events. In this paper, we exhibit the limitations of evaluating mitosis with AOGM using both simulated and real cell tracking data.

\end{abstract}


\section{INTRODUCTION}
\label{sec:intro}  

Multi-object tracking (MOT) is a computer vision task~\cite{ciaparrone2020deep}, whose aim is to detect and track objects (e.g., humans, cars, animals etc.) in a video. The process of MOT is consisted of two steps, the first step is recognizing variety of objects in different frames and extracting each features of it~\cite{liu2020deep} and the second step is to compare each object's feature and assign the corresponding object in the entire time-lapse video~\cite{duan2019centernet}. Through this way, we can obtain each object's entire mobile trajectory and their position information in a particular time. In biomedical research, cell tracking is a similar task to MOT, whose aim is to detect and track cells in a time-lapse microscopy video~\cite{ulman2017objective}. These two research fields are highly overlapped, all of them are about tracking objects in a continuous video but have some fundamental differences in terms of the different properties of the targeting objects.  

For instance, the MOT algorithms typically do not consider the object division, where a object is split two daughter objects, because typically most macroscopic objects like humans and cars will not be divided into two parts in a video. However, the cell division (mitosis) is prevalent and critical in microscopy videos and medical research~\cite{mitchison2001mitosis}. The evaluation method for MOT is not suitable for cell's tracking~\cite{mavska2014benchmark} for it doesn't consider object division. Therefore, acyclic oriented graphs matching (AOGM)~\cite{matula2015cell} which including cell's mitosis events has been widely accepted as the de facto standard measurement of evaluating the performance of cell tracking, rather than directly using the metrics in MOT, such as multi-object tracking accuracy (MOTA), ID Switches (IDS), ID F1 Score (IDF1) etc~\cite{milan2016mot16}.

However, based on our experiments, AOGM does not always function as expected for evaluating mitosis events, as shown in Figure 1. To get a good AOGM score (smaller means better in AOGM), the vertices (detected cells) and edges (links between cells) of the computed graphs are supposed to be almost the same as the ground truth. But as the limitation of video's quality or tracking algorithm, we cannot track mitosis event's procedure as exactly the same as the ground truth, and we realized that if the mitosis events were detected with small shifts, the AOGM scores were even worse than that if we totally ignored the mitosis events although we did detect the mitosis correctly. Therefore, the efforts of detecting mitosis might even lead to the inferior performance than just simply overlook the mitosis, which against the original intention of this method. In this study, we exhibit such phenomenons when evaluating mitosis using AOGM on both simulated and real cell tracking data. 

\begin{figure}[t]
\begin{center}
\includegraphics[width=1\textwidth]{{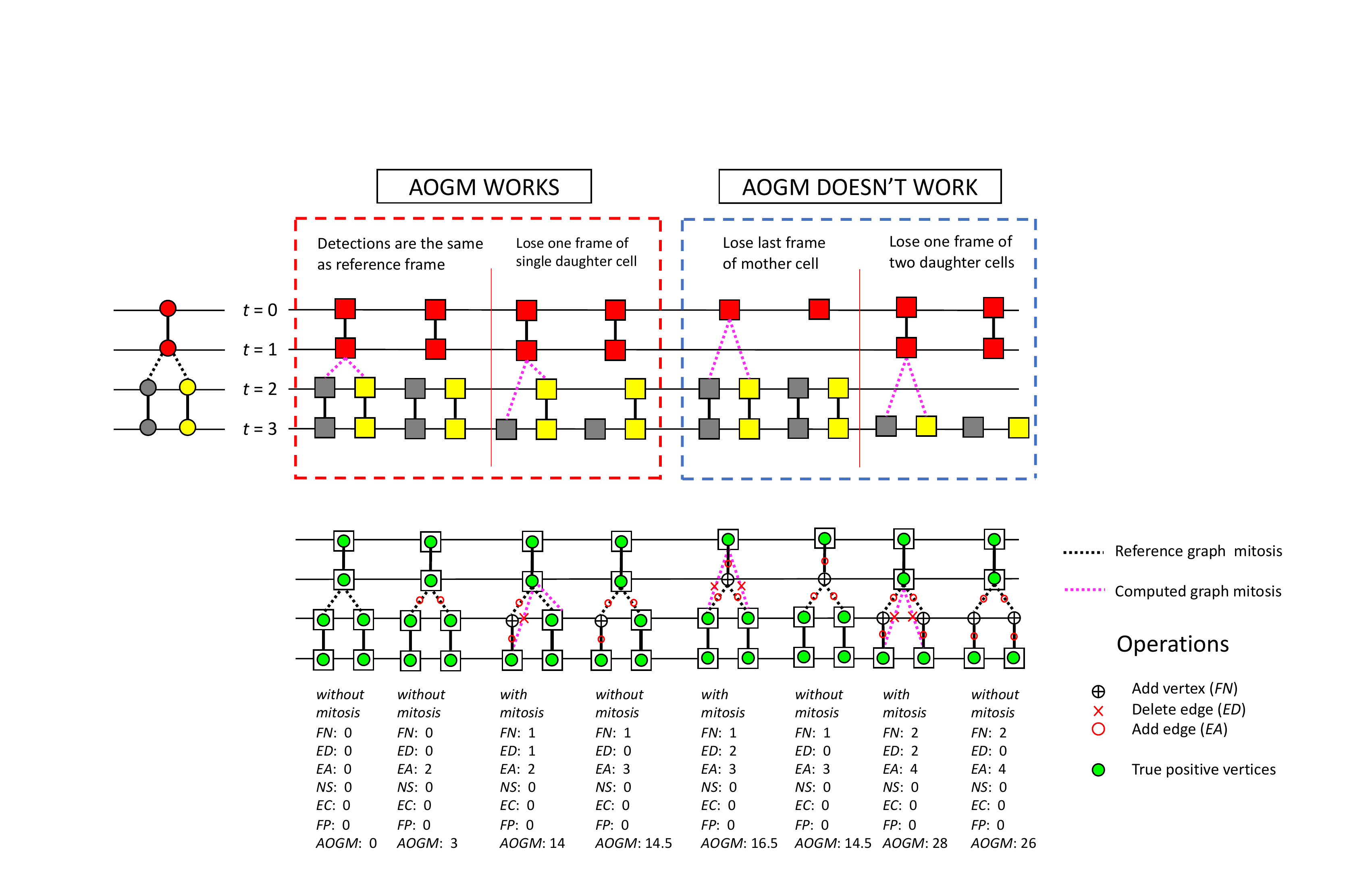}}
\end{center}
\caption{This figure shows the different mitosis scenarios and their corresponding AOGM measurements. AOGM is the weighted sum of the numbers of false negative vertices (\textit{FN}), false positive vertices (\textit{FP}) , missed splits (\textit{NS}), missing edges (\textit{EA}), redundant edges (\textit{ED}), and edges with the wrong semantics (\textit{EC}). Therefore, the algorithm with smaller AOGM is regarded as the one with better cell tracking performance. The colored circle vertices in left means reference cells and the two cases in the dashed box with square vertices means the prediction tracking we simulated. The blue dashed box indicates the scenarios that AOGM functions as expected, where the graphs with correct mitosis connection achieve smaller AOGM scores. By contrast, in the red dashed box, the AOGM scores are larger (worse) when connecting mother and daughter vertices, compared with just ignoring such mitosis events.} 
\label{Fig.1} 
\end{figure}

\section{Methods}
\subsection{AOGM}
AOGM is an extensively used cell tracking accuracy measure, which has continuously been used in different versions of Cell Tracking Challenge~\cite{ISBIchallenge2017}. AOGM is mathematically the weighted sum of the numbers of false negative vertices (\textit{FN}), false positive vertices (\textit{FP}) positives), missed splits (\textit{NS}), missing edges (\textit{EA}), redundant edges (\textit{ED}), and edges with the wrong semantics (\textit{EC}). The formula to calculate AOGM is 
\begin{equation}
A O G M=w_{N S} N S+w_{F N} F N+w_{F P} F P+w_{E D} E D+w_{E A} E A+w_{E C} E C .
\end{equation}
\noindent where $w_{NS}=5$, $w_{FN}=10$, $w_{FP}=1$, $w_{ED}=1$, $w_{EA}=1.5$, and $w_{EC}=1.5$, based on~\cite{matula2015cell}.

The examples of calculating AOGM for mitosis events are presented in Figure 1. The two cases in the blue dashed box indicate the scenarios that AOGM functions as expected, while the graphs in the red dashed box present the problematic cases that we discovered. The reference graph with colored circles indicates the mitosis event of a single cell from t0 to t3, and the four cases in the dashed box with square vertices shows different simulated prediction results of mitosis tracking. In each case, we divide the situation into two parts, the right part is the result we add the mitosis links between mother cell and daughter cell manually, and the left part is the same result without the linkage. We can assume all of these four cases with pink mitosis links track the mitosis event successfully for they do find a single cell's division, but the AOGM scores present different outcomes. For Case 3 and 4, the AOGM values of linking mother and daughter vertices are larger than without considering mitosis, which means although we detect the mitosis successfully we still will obtain worse scores on AOGM. For the cases in the red dashed box, we believe the graphs with larger AOGM are more legitimate as the cell-level consistent mitosis events are detected, compared with just ignoring the links between mother and daughter vertices.

\begin{figure}[ht]
\begin{center}
\includegraphics[width=1\textwidth]{{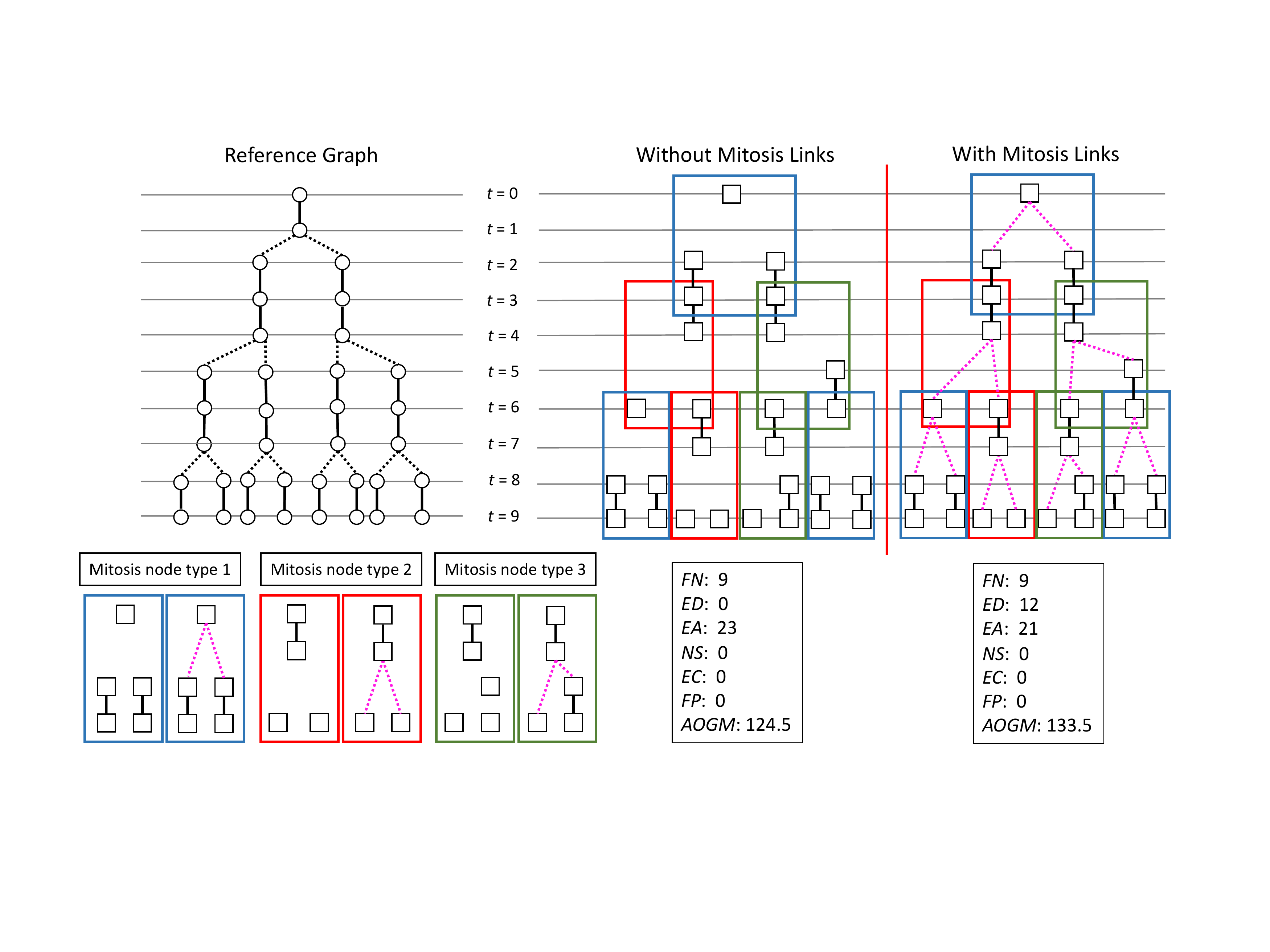}}
\end{center}
\caption{This figure presents the simulation to exhibit the limitation of AOGM when assessing cell mitosis. We simulated a reference graph with 7 mitosis events from t0 to t9. Then, two computed graphs were simulated with the same numbers vertices (squares). The left computed graph shows the case that we don't consider any mitosis, while the right computed graph shows the case that the graphs were linked with cell-level mitosis events. From the AOGM calculation, the left computed graph (without mitosis links) had smaller (better) AOGM scores than the right graph (with mitosis links). As a smaller AOGM indicates a better cell tracking results by definition, it shows that the tracking result without considering mitosis achieves the superior AOGM than the result with cell-level consistent mitosis detection.} 
\label{Fig.2} 
\end{figure}

\subsection{Simulation}
Figure 2 presents the simulation to exhibit the limitation of AOGM in assess cell mitosis. In order to present our result briefly, we simulate a situation with multiple mitosis node begin with a single cell. The reference graph with 7 mitosis events in the left panel presents different mitosis events from t0 to t9 and 8 daughter cells generated from the single cell at the beginning. Next, two computed graphs (predictions) are simulated with the same numbers of vertices (squares). 

We summarize three most common tracking mistakes which still can be considered as correct mitosis event detection in our daily experiment. Then, replace these mistakes mitosis node in simulated computed graph. There are three node type 1 and two node type 2, 3 in computed graph, which simulates several possible situations of detection results. After we arrange our results with removing linkage manually as shown in the left computed graph, we can find it clearly that the outcome of AOGM is less than the right one with correct linkage. The left computed graph shows the example that we don't track any mitosis, while the right computed graph demonstrates the example that the graphs are linked with correct cell-level mitosis events. 

As the vertices are kept the same between two simulated predictions, the \textit{FN}, \textit{FP} and \textit{NS} are identical for computing AOGM. The differences are from the \textit{ED} and \textit{EA}. We assume the right computed graph tracks should have a superior evaluation score for it not only has the same detection vertices as the left one but has the advantage that detecting mitosis events correctly. However, the outcome shows it has inferior(larger) AOGM value of 133.5. Through this simulation experiment, we can draw the conclusion that AOGM is not a good way to describe mitosis event as we may obtain worse outcomes even with correct mitosis detection.

\subsection{Empirical Validation}
To assess the limitation of AOGM in the real cell tracking data for mitosis evaluation, we employ the video Fluo-N2DH-GOWT1-01, which is one of simplest videos from ISBI Cell Tracking Challenge~\cite{ISBIchallenge2017}. It has 91 frames and 3 mitosis events in total, we screenshot the three frames and each of them has a particular cell that will have mitosis procedure in the next frame.

In order to observe the cells more precisely, we mark the cells which will have mitosis event with red box and record their entire mitosis process in timeline. The three full size pictures in the Figure 3 is frame 12, 42 and 73. It's apparent that each cell's position and appearance in each frame is distinct and some of them die away as time varied. In order to present the mitosis procedure of cells, we enlarge the picture of divided cells and put them in the time line so that we can fully analyze the procedure. 

We present the data set's ground truth in the upper of each case, and the dashed line presents the mitosis relationship between cells. From the graph, we have the knowledge that even though mother cell died in a particular frame, its daughter cell may not generate immediately in the sequence frame. On the contrary, the different daughter cell even have the same ancestors may appear in different frames, which brings obstacles to our tracking process. It also has some specific examples in video, like the cell's mitosis in third image only has one daughter cell. What's more, under some certain circumstances two cells may have collision and then merge to one cell, which will not be considered here. 

In this paper, we implement a single MOT baseline algorithm, called FairMOT~\cite{zhan2020simple}, to perform the object detection and cell tracking. We change the standard ground truth's format in ISBI from segmentation to FairMOT applied bounding box and train the model through our reorganized data source. After the training process, the model can track cell's trajectory as intended. As FairMOT is not designed for cell tracking, the cells are detected without considering the mitosis links. Then, we performed another simple linkage approach to link the detected mother cells and daughter cells as mitosis events. Briefly, if one or two cells with new tracks appear within five frames of another disappeared cell, and the center points of daughter cells located near the center of mother cells, those are linked as a mitosis event. After we obtain all results of mitosis, we computed the AOGM scores with/without mitosis links, which still indicates the limitation of AOGM evaluation, the results contain correct linkage has inferior outcomes.

\begin{figure}[ht]
\begin{center}
\includegraphics[width=1\textwidth]{{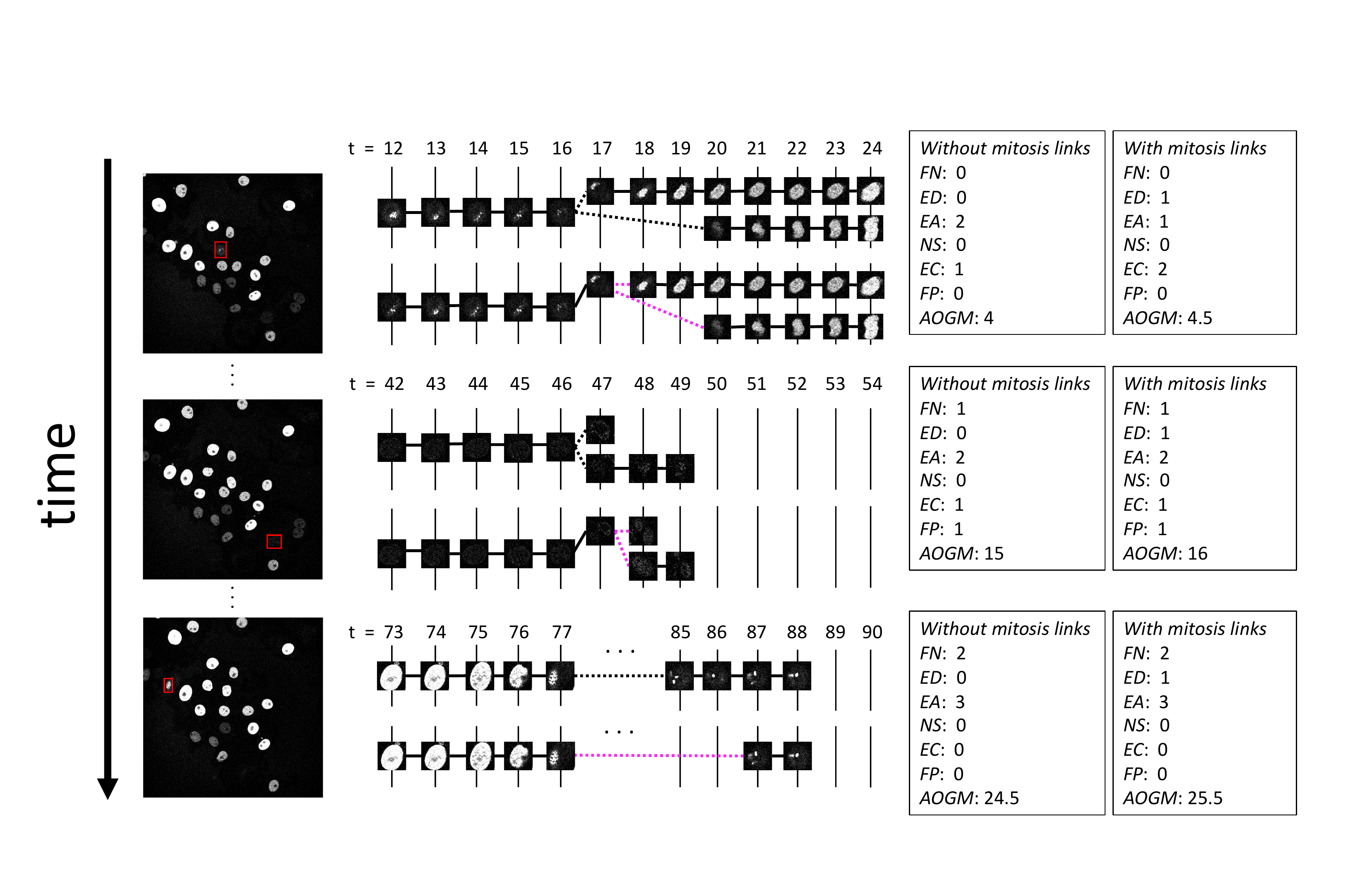}}
\end{center}
\caption{This figure presents the tracking results of all three mitosis events in the Fluo-N2DH-GOWT1-01 video from ISBI cell tracking challenge~\cite{ISBIchallenge2017}. We presented the results of using FairMOT~\cite{zhan2020simple}, which naturally did not model any mitosis (without mitosis links), and the same results but with mitosis links using post-processing. The AOGM scores are larger (worse) when modeling mitosis events than just ignoring the mitosis edges.} 
\label{Fig.2} 
\end{figure}

\section{Results}
From the simulation in Figure 2, the left computed graph (without mitosis links) has smaller AOGM than the right graph (with mitosis links). As smaller AOGM indicates a better cell tracking results by definition, it shows the problem that the tracking result without considering mitosis achieves superior AOGM than the other tracking result with cell-level consistent mitosis detection.

The empirical validation results are presented in Figure 3, where all three mitosis events from the Fluo-N2DH-GOWT1-01 video are presented. From the results, the direct detection and tracking results from FairMOT~\cite{zhan2020simple} (without considering mitosis) achieves smaller (better) AOGM than the results of linking mitosis events.

From the simulation and empirical validation above we can have the conclusion that AOGM is not always the best method to evaluate cell tracking because the weight of some particular correct mitosis detections will not be counted in the final score.

\section{Conclusions}  
In this paper, we exhibit the limitations of using AOGM to evaluate the performance of mitosis detection for cell tracking applications. As AOGM requires the exact same computed graphs as reference graphs to achieve superior performance, the results of cell-level consistent mitosis detection can even yield inferior performance compared with ignoring all mitosis. Therefore, an modified AOGM or other complementary measures (e.g., recall and precision of mitosis)~\cite{powers2020evaluation} might be needed to evaluate the cell tracking performance with mitosis.

\section{Acknowledgement}  
We acknowledge the help from Bryan Millis from Biomedical Photonics Center at Vanderbilt University, and Matthew Tyska from Department of Biology at Vanderbilt University.

\bibliographystyle{spiebib} 
\bibliography{paper}

\end{document}